\documentclass[letterpaper, 10 pt, conference]{ieeeconf}  

\IEEEoverridecommandlockouts                              

\usepackage{graphics} 
\usepackage{epsfig} 
\usepackage{mathptmx} 
\usepackage{times} 
\usepackage{amsmath} 
\usepackage{amssymb}  
\usepackage{subcaption}
\usepackage{tabularx} 
\usepackage{threeparttable}
\usepackage{gensymb} 
\usepackage{xcolor}
\usepackage{hyperref}
\hypersetup{
    colorlinks=true,
    urlcolor=blue,
    linkcolor=black,
    citecolor=black,
}

\title{\LARGE \bf
RflyUT-Sim: A Simulation Platform for Development and Testing of Complex Low-Altitude Traffic Control
}

\author{
Zonghan Li, Tianwen Tao, Rao Fu, Liang Wang, Dongyuan Zhang, Quan Quan
\thanks{Zonghan Li is with the School of Automation Science and Electrical Engineering, Beihang University, Beijing, 100191, China (Email: 13100075256@163.com)}
\thanks{Tianwen Tao is with Beijing Intelligent Token Technology Co., Ltd., Beijing, 100191, China (Email: taotevin@gmail.com)}
\thanks{Rao Fu is with the School of Traffic and Transportation, Beijing Jiaotong University, Beijing, 100044, China (Email: raofu@bjtu.edu.cn).}
\thanks{Liang Wang is with the School of Artificial Intelligence and Automation, Huazhong University of Science and Technology, Wuhan, 430070, China (Email: wangliang.f@gmail.com)}
\thanks{Dongyuan Zhang is with the School of Automation Science and Electrical Engineering, Beihang University, Beijing, 100191, China (Email: zhangdongyuan@buaa.edu.cn)}
\thanks{Quan Quan (Corresponding author) is with the School of Automation
Science and Electrical Engineering, Beihang University, Beijing, 100191,
China (Email: qq\_buaa@buaa.edu.cn)}
}

\begin{document}

\maketitle

\begin{abstract}

Significant challenges are posed by simulation and testing in the field of low-altitude unmanned aerial vehicle (UAV) traffic due to the high costs associated with large-scale UAV testing and the complexity of establishing low-altitude traffic test scenarios. Stringent safety requirements make high fidelity one of the key metrics for simulation platforms. Despite advancements in simulation platforms for low-altitude UAVs, there is still a shortage of platforms that feature rich traffic scenarios, high-precision UAV and scenario simulators, and comprehensive testing capabilities for low-altitude traffic. Therefore, this paper introduces an integrated high-fidelity simulation platform for low-altitude UAV traffic. This platform simulates all components of the UAV traffic network, including the control system, the traffic management system, the UAV system, the communication network , the anomaly and fault modules, etc. Furthermore, it integrates RflySim/AirSim and Unreal Engine 5 to develop full-state models of UAVs and 3D maps that model the real world using the oblique photogrammetry technique. Additionally, the platform offers a wide range of interfaces, and all models and scenarios can be customized with a high degree of flexibility. The platform's source code has been released, making it easier to conduct research related to low-altitude traffic. 

\end{abstract}

\section{INTRODUCTION}

With the rapid advancement of unmanned aerial vehicles (UAVs), low-altitude UAV traffic has become increasingly popular in recent years.
According to data from the official website of the Federal Aviation Administration (FAA), as of December 17, 2024, there are more than one million UAVs lawfully registered with the FAA in the United States \cite{FAA}.
In China, the low-altitude economy reached a scale of 500 billion yuan in 2023 \cite{China}.
China plans to expand the scale of its low-altitude economy— including drones, vertical take-off and landing aircraft and other general aviation components— to 1.5 trillion yuan (US\$207 billion) by the end of 2025\cite{2025The}. 
Numerous universities and institutions have initiated research in the field of low-altitude UAV traffic. 

However, real-world trials require large numbers of UAVs and extensive areas, which results in high costs and significant safety risks. 
Therefore, there is a significant demand for a high-precision simulation and assessment system designed explicitly for low-altitude UAV traffic scenarios.
The challenge of simulating low-altitude traffic lies in maintaining the fidelity of both UAV models and scene models while providing a comprehensive interface for testing traffic management simulations.
As illustrated in Table I, a comparison is presented between RflyUT-Sim and other analogous low-altitude UAV simulation platforms.
The Gazebo robot simulation platform, for example, is a widely used tool for simulating, developing, and testing UAVs. It features a robust physics simulation engine and sensor model library. However, Gazebo's simulation scenes lack realism, and it lacks development interfaces and tools tailored for large-scale traffic scenarios \cite{gazebo}. 

To facilitate the research of low-altitude UAV traffic, this paper establishes the RflyUT-Sim simulation platform.
The platform's advantages lie in the following aspects.

\begin{figure}[t]
	\centering
	\includegraphics[width=0.45\textwidth,height=0.25\textwidth]{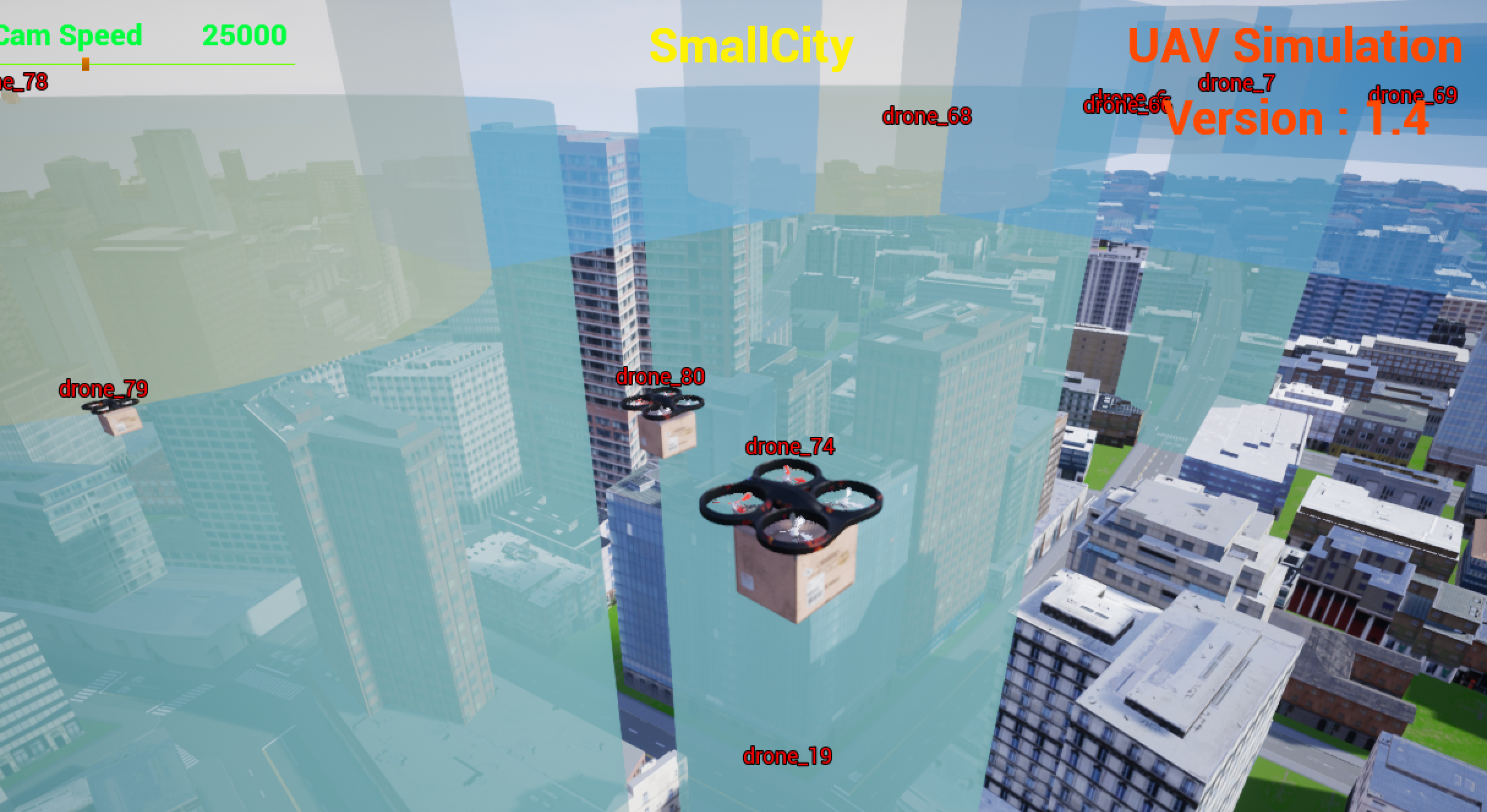}
	\caption{RflyUT-Sim simulation view}
	\label{simulation}
\end{figure}

\subsubsection{High-fidelity}

The primary advantage of RflyUT-Sim is attributable to its utilisation of a full-state model for the UAV, and an oblique photogrammetry modelling technique. It achieves high fidelity in terms of the UAV, sensors, and simulation scenarios. 
The calculations in RflyUT-Sim are precise, for the high-precision mathematical models can truly restore the simulation process of a low-altitude UAV traffic network.

\subsubsection{Comprehensive interfaces}

RflyUT-Sim offers a comprehensive set of simulation interfaces for transportation applications, including configuring airway network structures, testing algorithms, and managing logic. 

\subsubsection{Compatibility}

RflyUT-Sim utilises a modular design, whereby each independent module can be customised by the user, such as the UAV models and scenarios. 
Information exchange during simulation processes utilizes Redis\footnote{Redis (Remote Dictionary Server) is an open-source, in-memory data structure store used as a database, cache, and message broker, renowned for its high performance and support for diverse data types.} as a message bus, enabling compatibility with diverse systems. 

\newcolumntype{M}{>{\centering\arraybackslash}m{\dimexpr(0.9\textwidth-6\tabcolsep-6\arrayrulewidth)/7}} 

\begin{table*}[t]
	\caption{A comparison between RflyUT-Sim and other platforms}
	\label{table_comparison}
	\centering
        \renewcommand{\arraystretch}{1.5}
        \begin{threeparttable}
	\begin{tabular}{|M||M||M||M||M||M|}
		\hline
		Simulation platform & FlightGear\tnote{2} & Gazebo\tnote{3} & AirSim & RflySim & \textbf{RflyUT-Sim} \\
		\hline
		Hardware-in-the-Loop & Not supported & Supported & Supported & Supported & \textbf{Supported} \\
		\hline
		Sensor simulation & Not supported & Supported & Supported & Supported & \textbf{Supported} \\
		\hline
		Scene realism & Low & Medium & High & High & \textbf{High} \\
		\hline
		Airway network structure & Not available & Not available & Not available & Not available & \textbf{Available} \\
		\hline
		Traffic management testing capability & None & None & None & None & \textbf{High} \\
		\hline
	\end{tabular}
        \begin{tablenotes}
            \small
            \item[2] An overview of FlightGear can be found in \cite{FlightGear}.
            \item[3] An overview of Gazebo can be found in \cite{gazebo}.
        \end{tablenotes}
        \end{threeparttable}
\end{table*}

The outline of this paper is as follows. Section II presents the framework, which contains the architecture, realization, and interfaces of RflyUT-Sim. Section III provides a detailed introduction to the models of the RflyUT-Sim. The functions and performance are demonstrated in Section IV. In Section V, two demos are presented on the RflyUT-Sim platform. Finally, concluding remarks are stated in Section VI.

\section{RFLYUT-SIM FRAMEWORK}

\subsection{Architecture}

As shown in Fig. 2, RflyUT-Sim consists of four layers: the service gateway layer, the service layer, the engine layer, and the virtualized resource layer. The advantage of this architectural design is the segregation of user requests, services, simulation engines, and stored data. This segregation facilitates the independent development and management of each component, without the need for mutual interference.

\begin{figure}[b]
	\centering
	\includegraphics[scale=0.17]{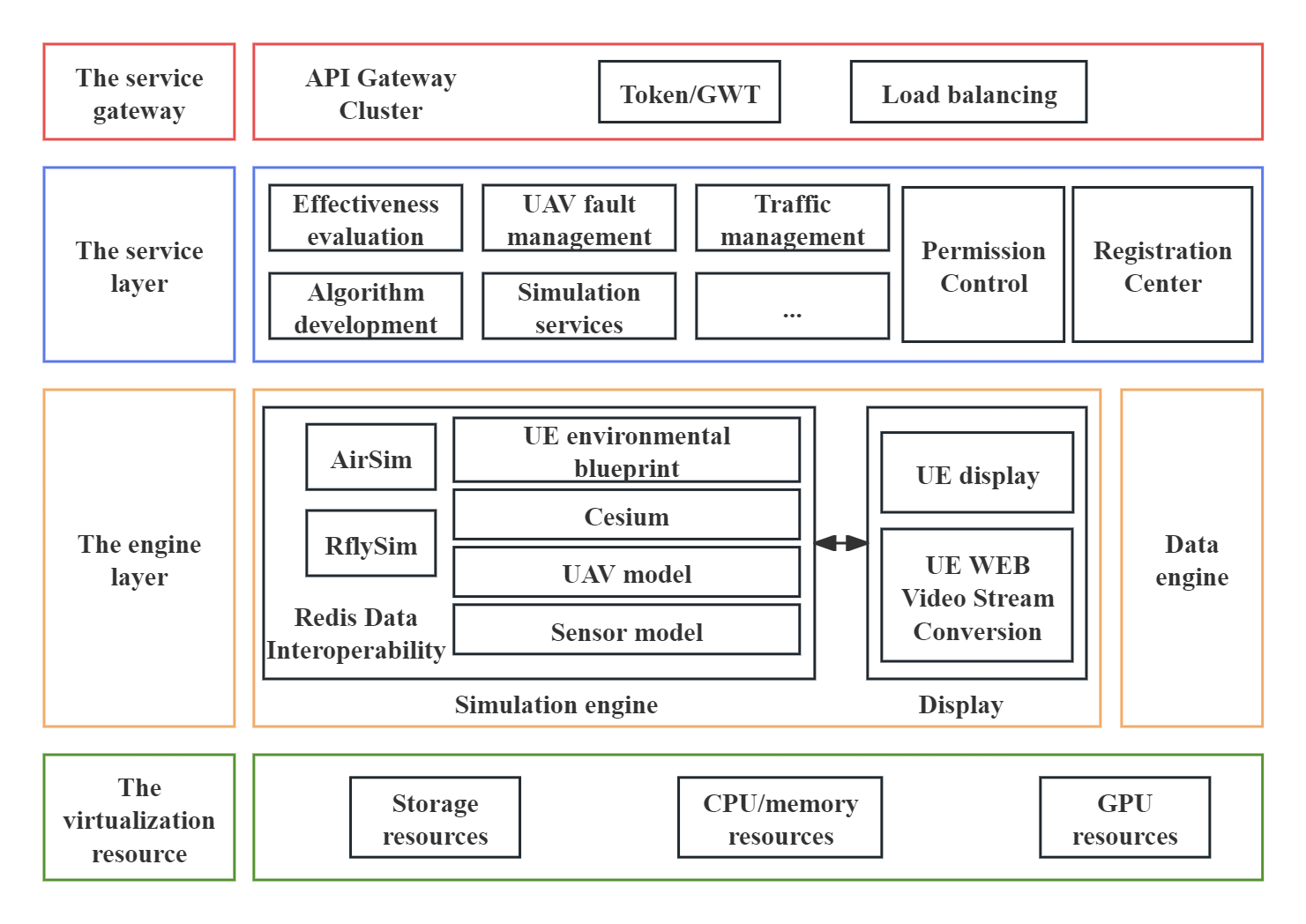}
	\caption{Four layers of RflyUT-Sim}
	\label{fig1}
\end{figure}

\subsubsection{The service gateway layer}

A user interface, standardised APIs, and stable network services are provided in the service gateway layer \cite{gateway2}. A service gateway performs the following functions: routing client requests to the correct microservices, implementing load balancing between clients and microservices, handling authentication, and managing different versions of APIs \cite{gateway}.

\subsubsection{The service layer}

The system encompasses the following services: effectiveness evaluation, algorithm development, UAV fault management, simulation services, and traffic management, among others. 

\subsubsection{The engine layer}

The engine layer consists of a simulation engine, a 3D rendering engine, and a data engine. 

The RflyUT-Sim system utilizes RflySim\cite{RflySim}/AirSim\cite{AirSim} as its simulation engine, which includes UAV models and sensor models.

The 3D rendering engine of RflyUT-Sim is Unreal Engine 5 (UE5) \cite{UE5}, which can render 3D models of UAVs, buildings, terrain, weather, and pedestrians, among other objects. Cesium\cite{Cesium} for Unreal Engine is used to import oblique photogrammetry data and create 3D maps of the real world.

The data engine can manage and process the data produced during simulation. 

\subsubsection{The virtualized resource layer}

The virtualization resource layer includes storage resources, CPU/memory resources, and GPU resources.

\subsection{Realization}

The RflyUT-Sim simulator is designed based on the UAV traffic management system framework \cite{Kopardekar2016UTM}, employing a distributed architecture. As shown in Fig. 3, the relationships among the subsystems are clearly described. The control subsystem and the traffic management subsystem are the systems under test. 

\begin{figure*}[htpb]
	\centering
	\includegraphics[scale=0.18]{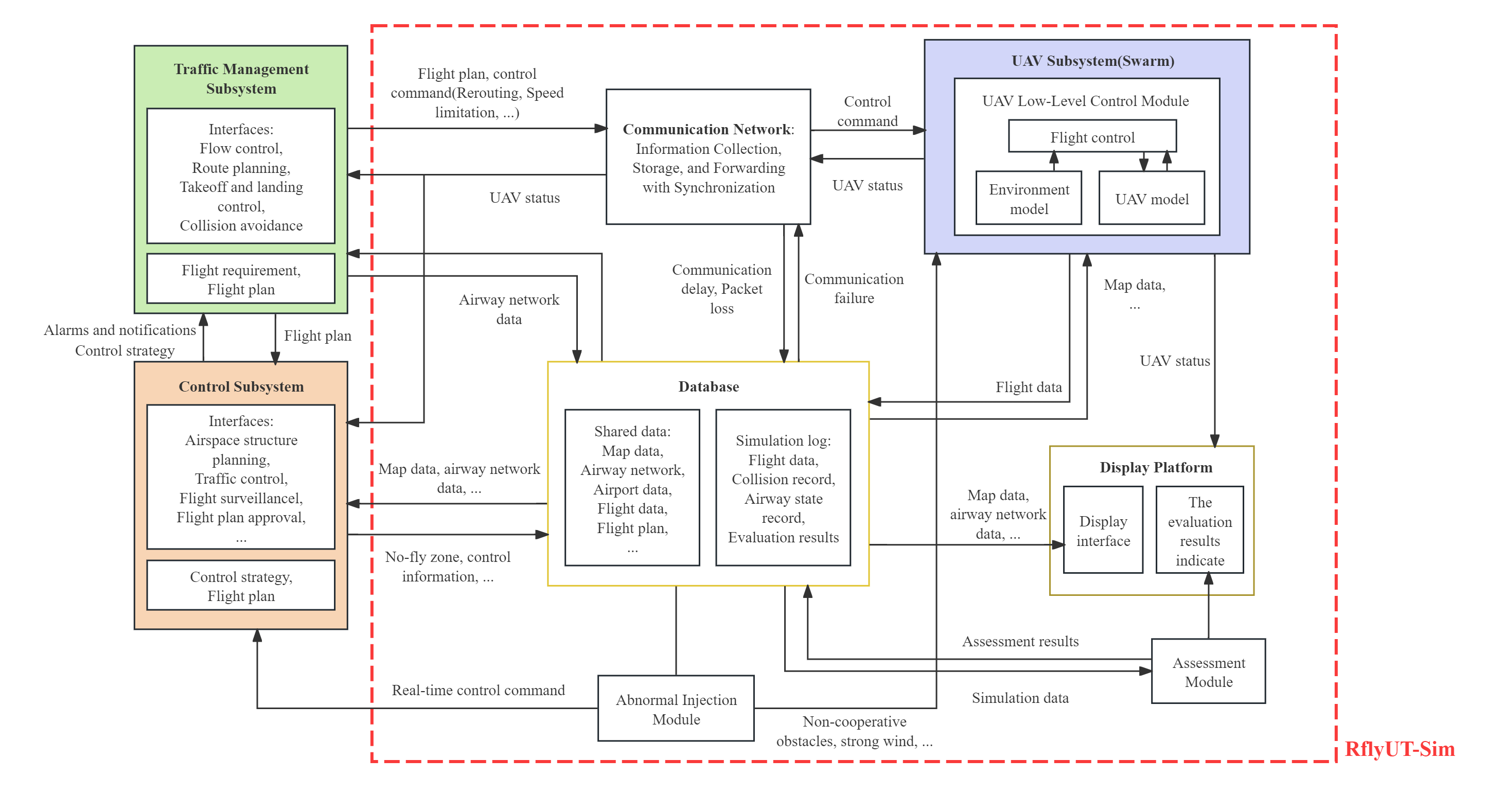}
	\caption{RflyUT-Sim data flow diagram}
	\label{fig2}
\end{figure*}

\subsubsection{The control subsystem}

The simulation process of the control subsystem is as follows.

\begin{itemize}
	
\item Initialization: The control system reads the airway and airport configuration parameters from the simulation proposal. It generates an airway network based on the loaded route structure rules, or directly loads the data of the airway network structure. The map data and airway network data are stored in a shared database. Additionally, the control strategy, the no-fly zone, and other relevant information are read and stored in the local database. 
\item During simulation: The control system is responsible for airspace structure planning, flight plan approval (including authorisation for take-off and landing), flight monitoring, and airway traffic statistics.
\item After simulation: The simulation log and other simulation data are stored in the system database. And the simulation cache data is cleared. 

\end{itemize}

\subsubsection{The traffic management subsystem}

The simulation process of the traffic management subsystem is as follows.

\begin{itemize}
	
\item Initialization: The flight requirement data from the simulation proposal is read and stored in the local database of the traffic management system.
\item During simulation: The traffic management system is responsible for decision-making regarding flight plans and risk assessment. This encompasses the control of UAV takeoff and landing, route planning, and collision avoidance, among other tasks. It issues control commands to the UAV in accordance with the flight path rules specified by the control system.
\item After simulation: The simulation log and other simulation data are stored in the system database. And the simulation cache data is cleared. 
	
\end{itemize}

\subsubsection{The UAV subsystem}

The UAV system is responsible for simulating UAV swarms within the air traffic system. The underlying control module for the UAV is implemented by RflySim or AirSim, simulating the communication network between the UAV swarm and individual UAVs. The UAV simulation management module is responsible for receiving information regarding the commencement and cessation of flight plans, and for generating or reclaiming corresponding simulation threads. This process serves to reduce overhead from inactive UAV threads. Furthermore, this module is responsible for receiving control commands, distributing them to UAVs with matching IDs, and forwarding messages transmitted by the UAVs \cite{AirSimUAV}.

The UAV system provides low-level control interfaces for flight control, obstacle avoidance algorithms, and visual perception. 

\subsubsection{The communication network}

The communication network is responsible for collecting, storing, forwarding, and synchronizing information among system modules. The communication module simulates factors such as packet loss and congestion that are found in real-world physical communication networks.

\subsubsection{Anomaly injection}

The Anomaly Injection Module is responsible for generating faults and anomalies within UAV traffic scenarios. Anomalies emerge from two distinct origins: initial anomaly injection schemes and real-time generation via the console. The following types of anomalies are included:

\begin{itemize}
	
\item Control: Airspace control, traffic control, etc.
\item Environment: Non-cooperative obstacles, strong winds.
\item UAV: Motor failure, propeller breakage, etc.
\item Communication Module: Communication delay, packet loss, etc.
	
\end{itemize}

\subsection{Interfaces}

The interfaces of RflyUT-Sim are primarily categorized into simulation control interfaces and simulation scenario interfaces.

\subsubsection{The simulation control interfaces}

The simulation control interfaces of RflyUT-Sim are shown in Fig. 4.

\begin{itemize}
	
\item UAV control: RflyUT-Sim provides interfaces for issuing control commands, acquiring position and velocity status, obtaining perception information, and fault injection.
\item Acquisition of airway network information: Users can access information related to airways, nodes, and airports, which are fundamental structures of an airway network, through interfaces. 
\item Flight plan interfaces: Interfaces related to flight plans include those for submitting flight plans and querying approval results.
\item Flight Surveillance: Users can access information related to the flight status of UAVs. 
\item Other interfaces such as those for acquisition of map information, flight Plan Approval, traffic/airspace control, etc.

\end{itemize}

\begin{figure}[t]
	\centering
	\includegraphics[scale=0.15]{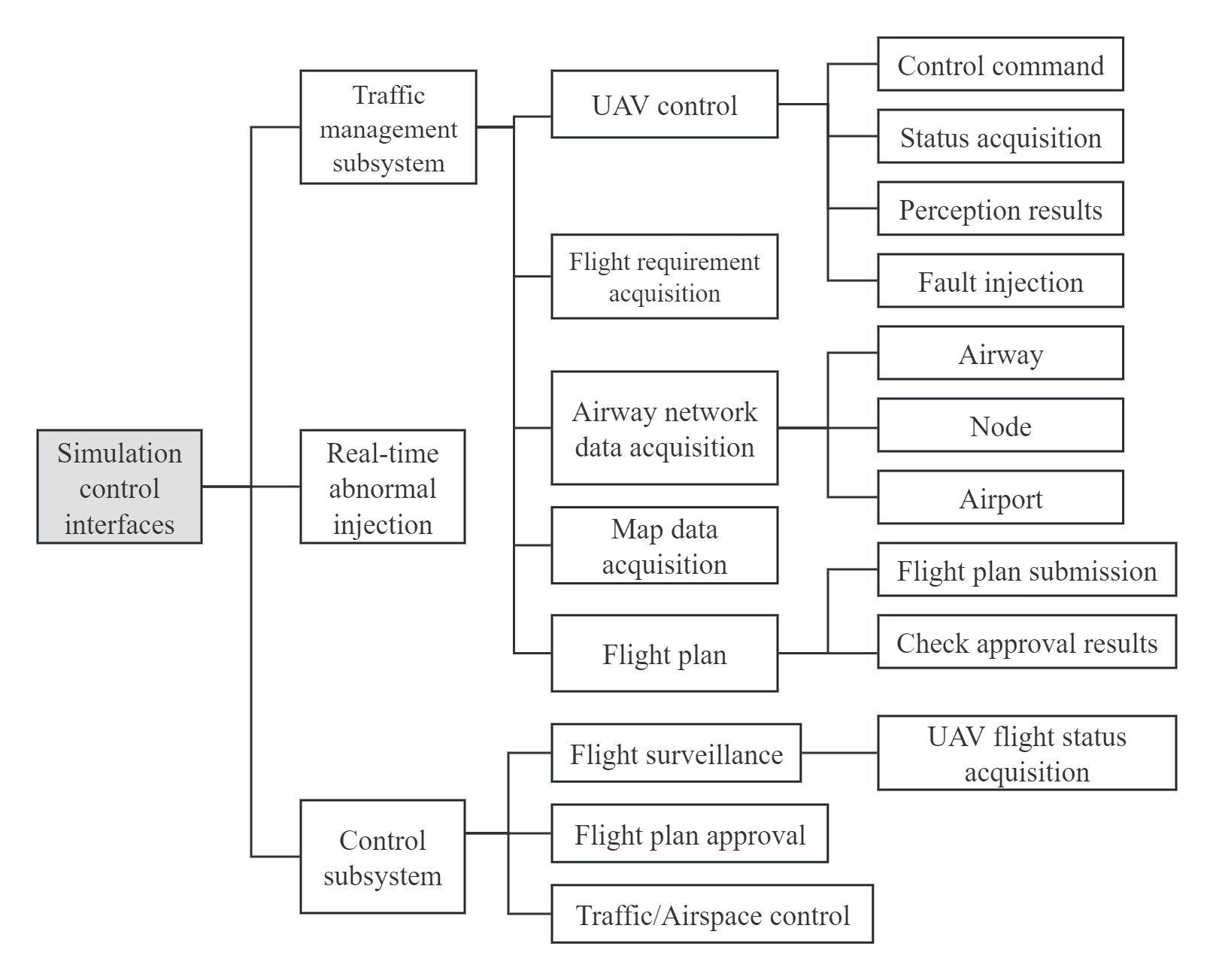}
	\caption{The simulation control interfaces of RflyUT-Sim}
	\label{fig3}
\end{figure}

\subsubsection{The simulation scenario interfaces}

The simulation scenario interfaces of RflyUT-Sim is shown in Fig. 5.

\begin{figure}[b]
	\centering
	\includegraphics[scale=0.13]{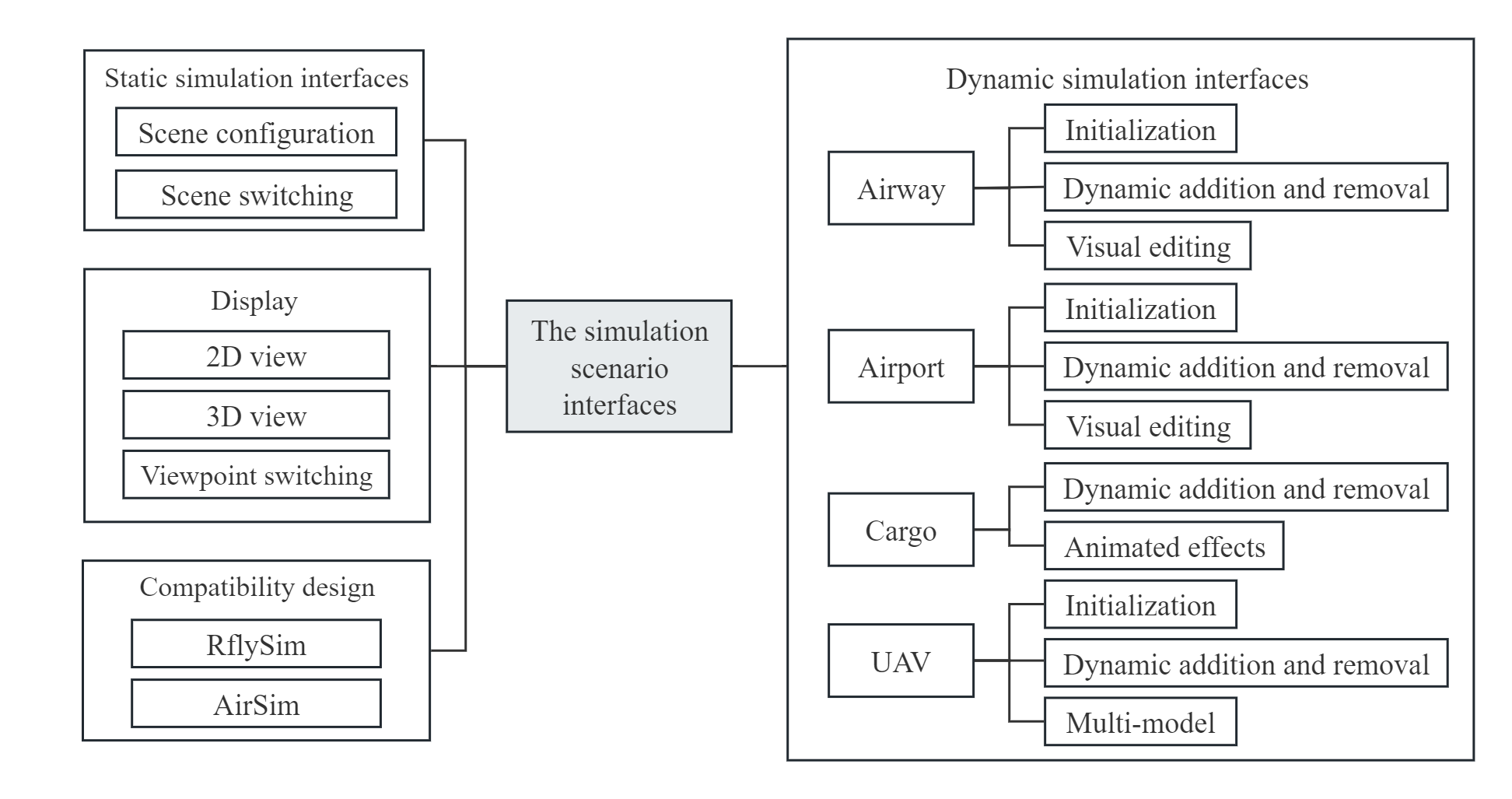}
	\caption{The simulation scenario interfaces of RflyUT-Sim}
	\label{fig4}
\end{figure}

\begin{itemize}
	
\item Dynamic simulation interfaces: The airway network structure, which encompasses airways, nodes, and airports, facilitates visual editing and manipulation. Animated cargo effects are displayed during the pickup and delivery phases of the process. Users have the capacity to customize UAV models. All scene entities above are equipped with interfaces that facilitate initialization and dynamic scaling. 
\item Static simulation interfaces: The static simulation interfaces include a scene configuration interface and a scene switching interface. Users can customize their own traffic scenarios according to their specific requirements.
\item Simulation display interfaces: Users have the capacity to access both the 3D simulation view and the simulation view within the 2D map. Furthermore, the viewpoint within the scene can be freely moved. 
\item Compatibility design: The simulation scenarios of RflyUT-Sim offer a framework for the effective operation of RflySim and AirSim interfaces. The importation of both RflySim and AirSim simulators is permitted. 
	
\end{itemize}

\subsection{Hardware-in-the-Loop Setup}

RflyUT-Sim enables Hardware-in-the-Loop (HIL) simulation by connecting the physical autopilot system (flight controller) to a real-time simulation computer. The flight controller and the motion simulation model communicate via a USB data line, using a serial protocol to exchange sensor data and control signals, enabling real-time interaction. This setup allows for the validation of the algorithm's performance in conditions that closely mimic real-world scenarios, while also testing the compatibility and reliability of the hardware \cite{HIL}. 

\section{SYSTEM MODELING}

In this section, the model of the RflyUT-Sim system is introduced in detail.

\subsection{Airway network model}

The data of the airway network are stored in the database, as shown in Fig. 3. The airway network structure employs a universal framework composed of airways, nodes, and airports. An airway is defined as a designated airspace for the operation of UAVs. A node is defined as a connection point between different airways. An airport is defined as a facility on the ground where UAVs take off and land. It connects to nodes within the airway network\cite{network}.

Users can configure the data of the above airway network structure themselves. The system also provides interfaces for establishing pipelines, allowing users to define airspace structures according to their needs freely. 

\begin{figure}[h]
	\centering
	\begin{subfigure}[t]{0.40\textwidth}
		\includegraphics[width=\textwidth,height=0.45\textwidth]{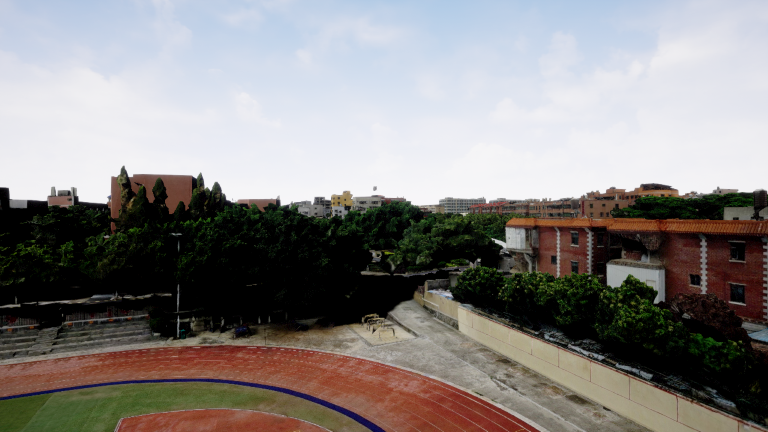}
		\caption{RGB camera}
		\label{fig7.1}
	\end{subfigure}
	
	\begin{subfigure}[b]{0.40\textwidth}
		\includegraphics[width=\textwidth,height=0.45\textwidth]{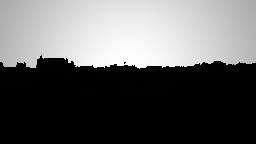}
		\caption{Depth camera}
		\label{fig7.3}
	\end{subfigure}
	
	\begin{subfigure}[b]{0.40\textwidth}
		\includegraphics[width=\textwidth,height=0.45\textwidth]{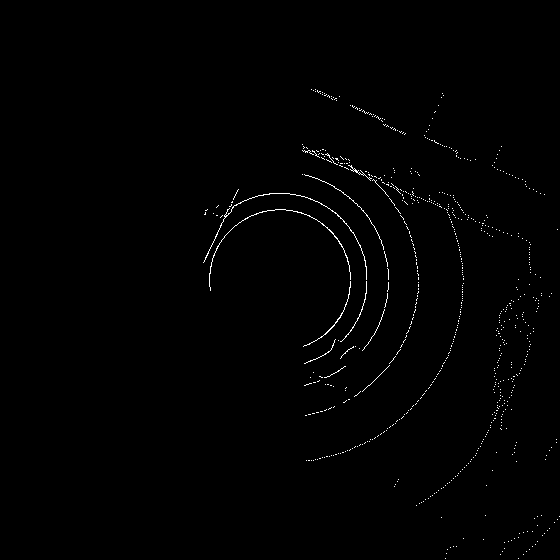}
		\caption{LiDAR}
		\label{fig7.4}
	\end{subfigure}
	\caption{Different sensors}
\end{figure}

\subsection{UAV model}

The RflyUT-Sim system supports modeling of various types of UAVs, based on the RflySim/AirSim engine. The following categories of UAVs are commonly modeled: fixed-wing, multi-rotor, and vertical takeoff and landing (VTOL) types, which include eVTOL types.

The UAV model under consideration employs kinematic and dynamic modeling to simulate the physical characteristics of real UAVs \cite{AirSim2021Physics}.

Most parameters of UAV are editable, such as mass, gravitational acceleration, rotational inertia matrix, thrust coefficient, torque coefficient, motor response time constant, drag coefficient, damping torque coefficient, etc. 
Consequently, users are free to import their own UAV models for testing purposes. 

Simultaneously, the 3D map model can simulate data from various sensors, such as visible light cameras, depth cameras, and LiDAR, as shown in Fig. 6, and transmit the data to the control unit.

\subsection{3D map model}

The RflySim/AirSim simulator's outputs are connected to the 3D renderer engine constructed using Unreal Engine 5. The renderer displays real-time simulation information, including the UAVs' position, attitude, and collision status, etc. 
Users can switch between viewpoints at any time: overall view, single UAV view. Alternatively, they can position the observation camera to a specific view as they prefer.  

The 3D map model facilitates the generation of virtual 3D maps of specific real-world regions, utilizing oblique photography technology \cite{qxsy} and customized to user specifications.
A 3D map based on the oblique photography data of an area in Xiamen, China, is shown in Fig. 7.

\begin{figure}[b]
	\centering
	\begin{subfigure}[b]{0.44\textwidth}
		\includegraphics[width=\textwidth,height=0.5\textwidth]{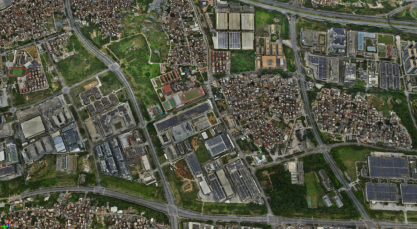}
		\caption{Bird's-eye View}
		\label{xiamen}
	\end{subfigure}

	\begin{subfigure}[b]{0.44\textwidth}
		\includegraphics[width=\textwidth,height=0.5\textwidth]{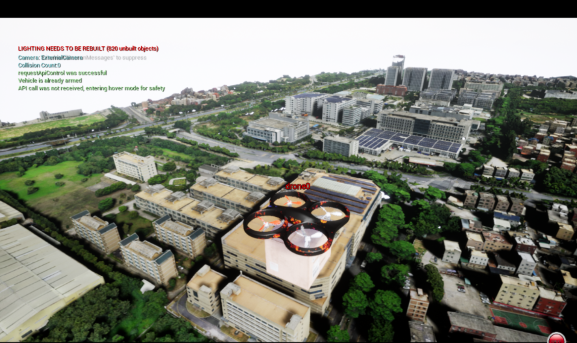}
		\caption{A single UAV perspective}
		\label{xiamen2}
	\end{subfigure}
	\caption{The 3D map of an area in Xiamen}
\end{figure}

\section{FUNCTION AND PERFORMANCE}

\subsection{Testing of the control system}

As demonstrated in Fig. 3, the control system can be replaced with the user's own system. 
The control system performs the following functions during its operation:

\begin{itemize}
	
\item Traffic control
\item Flight Plan Approval
\item Airway Traffic Statistics
\item UAV flight monitoring
	
\end{itemize}

The performance of the control system under test can be evaluated through analysis of the simulation results and data by establishing a connection between the control system and RflyUT-Sim.

\subsection{Testing of the traffic management system}

As demonstrated in Fig. 3, the traffic management system can be replaced with the user's own system. 
The traffic management system performs the following functions during its operation:

\begin{itemize}
	
\item Route planning
\item Flight Plan Decision-Making
\item Takeoff and Landing Control
\item Flight control
	
\end{itemize}

The performance of the traffic management system under test can be evaluated through analysis of the simulation results and data by establishing a connection between the traffic management system and RflyUT-Sim.

\subsection{Algorithm design} 

RflyUT-Sim provides users with interfaces for algorithm design. Users are at liberty to design their own algorithms and to optimize them through simulation testing. RflyUT-Sim is capable of evaluating a variety of algorithms, including the following ones\cite{pathplan,Park2023Vertiport}: 

\begin{itemize}
	
	\item Path planning algorithm
	\item UAV control algorithm
	\item Navigation algorithm
	\item Perception algorithm
	\item ...
	
\end{itemize}

\subsection{Hardware-in-the-Loop simulation} 

RflyUT-Sim has the capacity to facilitate the concurrent simulation of UAVs connected to hardware flight controllers and virtual UAVs within a single scenario. RflyUT-Sim utilizes virtual UAVs to simulate large-scale UAV traffic scenarios. UAVs equipped with hardware flight controllers execute tasks within these scenarios, enabling the testing and evaluation of the aforementioned hardware flight controllers in traffic conditions.

\subsection{Testing under abnormal conditions} 

In practical deployments, low-altitude traffic systems are exposed to various external disturbances. Bird activity poses a collision risk to UAVs and may disrupt their flight operations, while adverse weather conditions such as fog can significantly impair the sensing capabilities of onboard vision systems. Therefore, it is essential to incorporate various anomalous factors into the simulation process.

RflyUT-Sim incorporates anomaly injection interfaces, which are capable of generating anomalies, including severe weather, flying obstacles, and UAV malfunctions, during the simulation process. These interfaces can be used to assess the overall UAV traffic system's capacity to manage anomalies and its stability. 

\subsection{Data-driven algorithm training}

Due to the high cost and safety concerns associated with real-world UAV testing, flight data has become a valuable resource for unmanned aerial vehicles. Such data can be used as datasets for applications including UAV health monitoring and algorithm training. RflyUT-Sim generates a large amount of data during the simulation process. The simulation logs record comprehensive parameters of the UAV throughout its flight, including position, orientation, velocity, sensor data, and collision events. Thanks to the high-fidelity modeling of the UAV dynamics, the simulated data can serve as a reliable substitute for real flight data to a significant extent\cite{data}.

For instance, these trajectory data can be employed to train diffusion model-based algorithms for feasible trajectory generation. Each simulation run produces over one hundred flight trajectories, significantly improving the efficiency of algorithm training.

Furthermore, the simulation data generated by RflyUT-Sim can be utilized for data fusion and visualization within air traffic management systems. 

\subsection{Performance metrics} 

RflyUT-Sim's performance metrics are shown in Table II. The resolution and frame rate were obtained under test conditions involving one hundred UAVs.

\newcolumntype{M}[1]{>{\centering\arraybackslash}m{#1}}

\begin{table}[h]
	\caption{The performance metrics of RflyUT-Sim}
	\label{table_performance}
	\centering
        \renewcommand{\arraystretch}{1.8}
	\begin{tabular}{|M{3.5cm}||M{3.5cm}|} 
		\hline
		Maximum number of UAVs & $\geq$100 \\
		\hline
		Maximum resolution of the visual sensor & 1K \\
		\hline
		Frame rate & $\geq$30FPS \\
		\hline
		Accuracy of oblique photogrammetry maps & Planimetric Accuracy: $\leq$0.3m
		Vertical Accuracy: $\leq$0.8m \\
		\hline
	\end{tabular}
\end{table}


\section{DEMOS}

\subsection{Demo objectives} 

\subsubsection{Objective 1}

The primary objective of the first demo is to establish a low-altitude traffic network comprising 100 UAVs to accomplish logistics delivery tasks.
This demo focuses on testing whether the system operates normally and whether functional metrics are met when simulating a drone traffic network at the scale of hundreds. The simulation is conducted using a virtual urban map named \textit{SmallCity}.

\begin{figure}[t]
	\centering
	\begin{subfigure}[b]{0.44\textwidth}
		\includegraphics[width=\textwidth,height=0.5\textwidth]{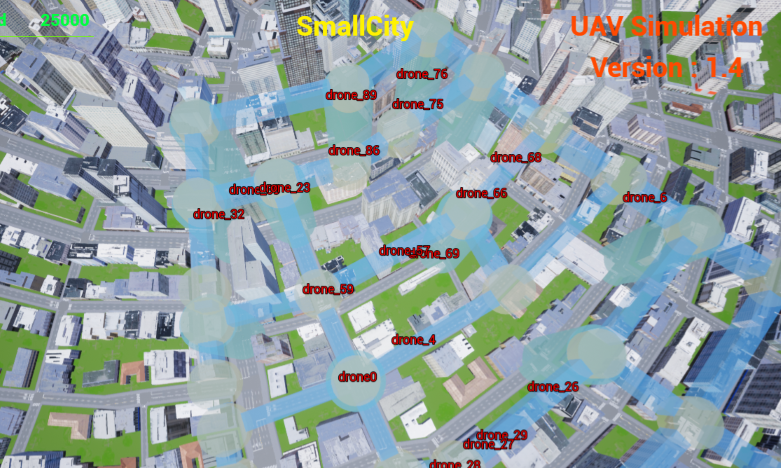}
		\caption{Bird's-eye View}
		\label{100}
	\end{subfigure}
	
	\begin{subfigure}[b]{0.44\textwidth}
		\includegraphics[width=\textwidth,height=0.5\textwidth]{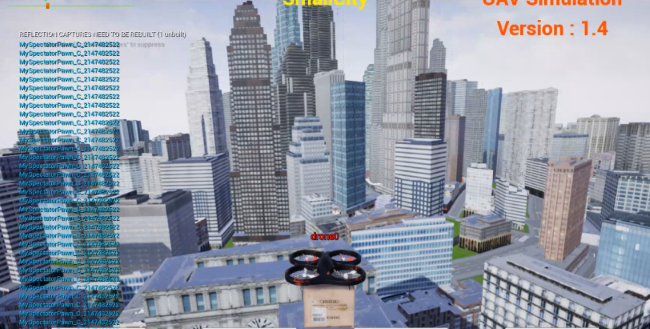}
		\caption{A single UAV perspective}
		\label{100_2}
	\end{subfigure}
	\caption{100 UAV simulation view on SmallCity map}
\end{figure}

\begin{figure}[t]
	\centering
	\includegraphics[width=0.44\textwidth,height=0.24\textwidth]{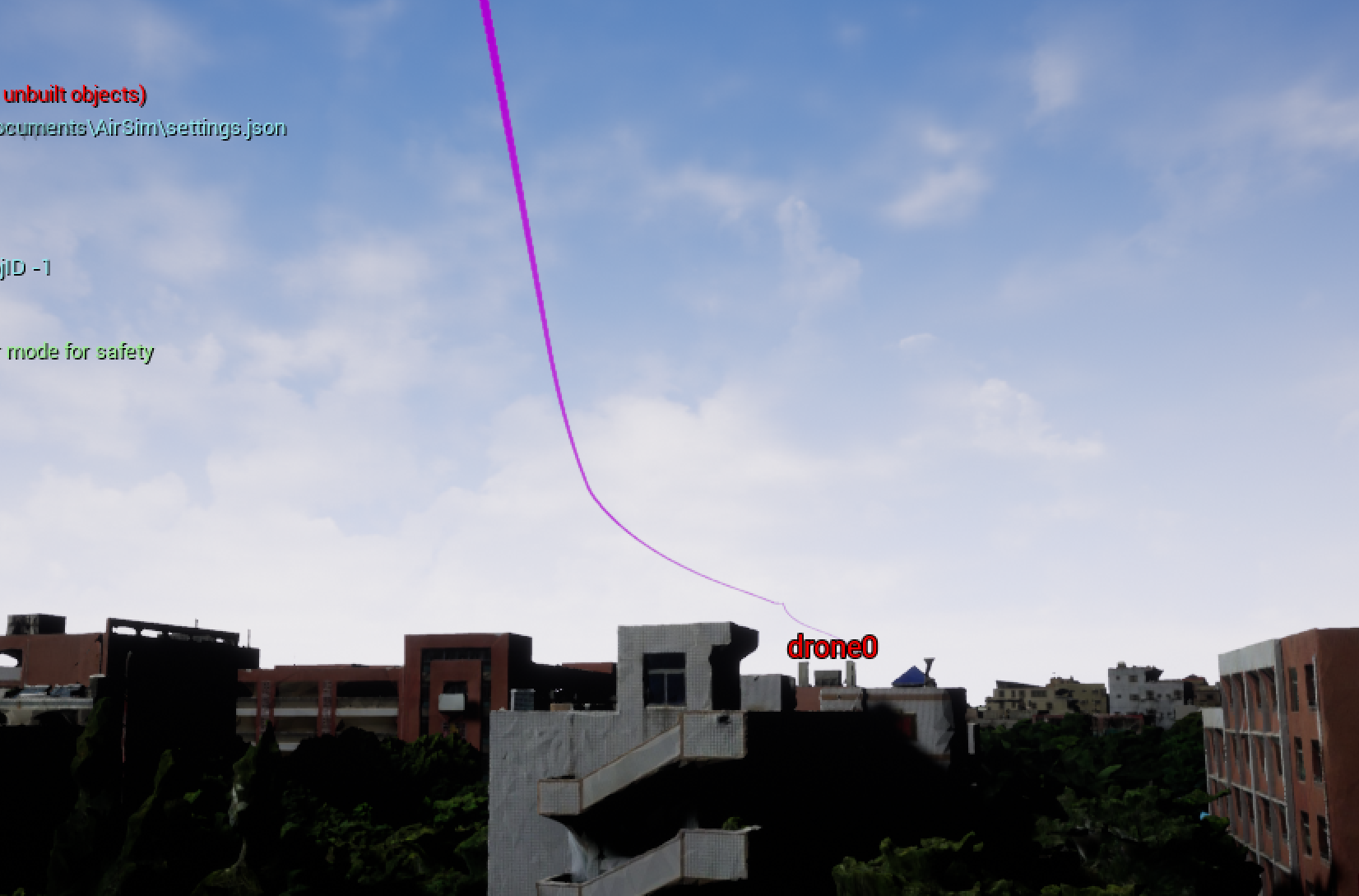}
	\caption{Flight path during obstacle avoidance on Xiamen map}
	\label{avo}
\end{figure}

\subsubsection{Objective 2}

The objective of the second demo is to achieve collision avoidance using LiDAR sensor data. 
The VFH algorithm\cite{Borenstein1991VFH} is adopted to identify obstacles by utilizing point cloud data acquired from the LiDAR sensor\cite{Glennie2020LiDAR}, to control the UAV to avoid the obstacles\cite{Wolcott2017Obstacle}. The simulation is conducted using a real-world map of a district in Xiamen, China.

\subsection{Implementation of two objectives} 

\subsubsection{Implementation of objective 1}

An airway network comprising 13 airports, 45 nodes, and 67 bidirectional airways is established. The altitude of this airway network is 120 meters. 
One hundred quadcopter UAVs and their corresponding flight requirements are configured and stored in the local database.

\subsubsection{Implementation of objective 2}

A flight route with a height of 18 meters and four sections featuring obstacles is designed.
The parameters of the LiDAR sensor are as follows: 12 channels, horizontal scanning angle 0\degree ~ 360\degree, vertical scanning angle -5\degree\string~5\degree, and a rotation speed of 100 revolutions per second. 

\subsection{Results} 

\subsubsection{Results of objective 1}

All 100 flight missions were successfully completed. None of the 100 UAVs collided. The simulation view of 100 UAVs on the SmallCity map is shown in Fig. 8.

\subsubsection{Results of objective 2}

The UAV successfully avoided obstacles throughout the entire flight without any collisions. The flight path during obstacle avoidance on the Xiamen map is illustrated in Fig. 9.  

\subsection{Source code and official website} 

The Python source code of RflyUT-Sim is published on GitHub: 

\begin{center}
\url{https://github.com/H-andle/RflyUT}
\end{center}

The official website contains demonstration videos and user guides for  RflyUT-Sim, which are available at the following address:

\begin{center}
\url{https://rflyut.com}
\end{center}

\hypersetup{urlcolor=black}

\section{CONCLUSIONS}

This paper proposes a simulation platform for low-altitude UAV traffic, which is of great assistance to scientific research and education. RflyUT-Sim is developed based on RflySim/AirSim, which is a UAV simulation engine that provides high-precision UAV simulation models. It also integrates Unreal Engine 5, a 3D rendering engine with rich simulation scenarios. The relationship among the subsystems and the usage of the interfaces is illustrated in detail. The simulation scenarios and models of RflyUT-Sim can be customized according to specific requirements, enabling RflyUT-Sim to serve a wide range of application scenarios.






\bibliographystyle{IEEEtran}
\bibliography{IEEEabrv,mybibfile}

\end{document}